\documentclass{ifacconf}

\usepackage{graphicx}      
\usepackage{natbib}        
\usepackage{algorithm}
\usepackage{algpseudocode}
\usepackage{amsmath}
\usepackage{amssymb}
\usepackage{subcaption}
\usepackage{placeins}
\usepackage{enumitem}
\usepackage{url} 
\usepackage{afterpage}

\makeatletter
\def\@logohead{%
\bgroup
\scriptsize
\savebox{\@tempboxa}{\vtop {%
\hbox to \textwidth{\hfill \copyright~2026 the authors. This work has been accepted to IFAC for publication under a Creative Commons Licence\hfill}%
\hbox to \textwidth{\hfill CC-BY-NC-ND.\hfill}%
\hbox to \textwidth{}%
\hbox to \textwidth{}%
\hbox to \textwidth{}%
}}%
\box\@tempboxa
\egroup
}
\makeatother

\begin{document}

\begin{frontmatter}

\title{Low-Resolution Perception for Robotic Packing} 

\author[First]{Giuseppe Fabio Preziosa} 
\author[First]{Federico Vignoni}
\author[First]{Chiara Castellano} 
\author[First]{Marco Faroni}
\author[First]{Andrea Maria Zanchettin}
\author[First]{Paolo Rocco}

\address[First]{The authors are with Politecnico di Milano, Piazza Leonardo da Vinci, 32.
Milano (Italy). e-mail: (giuseppefabio.preziosa, federico.vignoni, chiara2.castellano, marco.faroni, andreamaria.zanchettin, paolo.rocco)@polimi.it}


\begin{abstract}
This work tackles the problem of scalable perception for robotic packing with low-cost, low-resolution depth sensing. We propose a framework where reconstruction cues drive next-view selection and grasp evidence updates a per-object stability estimate, jointly deciding \emph{what} to acquire next and \emph{when} to grasp. During the reconstruction, a low-resolution Next Best View (NBV) strategy explicitly avoids redundant views while preserving task-relevant geometry. We validate the approach in two steps: (i) an ablation study of the utility function under very low resolution, and (ii) a full end-to-end evaluation across policies, showing how \emph{low-resolution} perception is a practical, scalable option for robotic packing.
\end{abstract}


\end{frontmatter}

\FloatBarrier
\section{Introduction}

Robotic packing lies at the intersection of flexible manufacturing and modern logistics. The task—detecting incoming items, estimating size, selecting and executing grasps, and placing objects into suitable containers—remains challenging despite advances in perception. While the workflow is consistent across applications, variability in object geometry and arrangement makes solutions broadly transferable. In practice, however, container selection and object arrangement still rely on tacit knowledge, limiting repeatability and consistency.
Prior work has largely focused on maximizing in-box spatial efficiency: \cite{Packing_just_ordering_offline} optimize placement under stability constraints, while \cite{Packing_just_ordering} extend this to both known and unknown objects, typically assuming high-resolution sensing and offline planning. More recent efforts target deployability, using minimalistic grippers (\cite{Packing_low_cost}) and RGB cameras (\cite{Packing_rgb_data}) to reduce system complexity. In this context, we investigate whether effective packing requires detailed perception or only task-relevant information, and how far perception can be reduced without compromising throughput and reliability.

Low-resolution sensing has proven effective in other domains (e.g., safety) \cite{spots}, but object understanding remains challenging due to limited detail and the need for multiple viewpoints. However, robotic packing does not require full 3D reconstruction: task-sufficient estimates of object dimensions are often enough for container selection and placement. Prior work shows that partial reconstructions (25–30\%) can support robust manipulation \cite{graspingPaper2,graspingPaper3}. The challenge becomes more pronounced in multi-object scenarios, where multiple acquisitions are needed and efficient decisions about which object to scan and when it is ready to pick are critical for throughput.
Traditional \emph{Next Best View} (NBV) strategies, designed for high-fidelity reconstruction, often lead to redundant poses and limited viewpoint diversity under low resolution. Moreover, dense but partial reconstructions can mislead grasp synthesis, producing grasps that fail on the full object. To address this, we extend our previous work \citet{preziosa2025low} by proposing an end-to-end perception–planning pipeline for multi-object robotic packing with low-cost, low-resolution depth sensors. Specifically, we integrate the LR--NBV strategy from \citet{preziosa2025low}, which reduces redundancy through density-aware exploration, within a framework that fuses grasp hypotheses to determine whether an object is ready for picking or requires further views. A video overview is available at \url{https://youtu.be/9uRI0bNqhDk}.

\begin{figure*}[!t]
\begin{center}
\includegraphics[width=16.5cm]{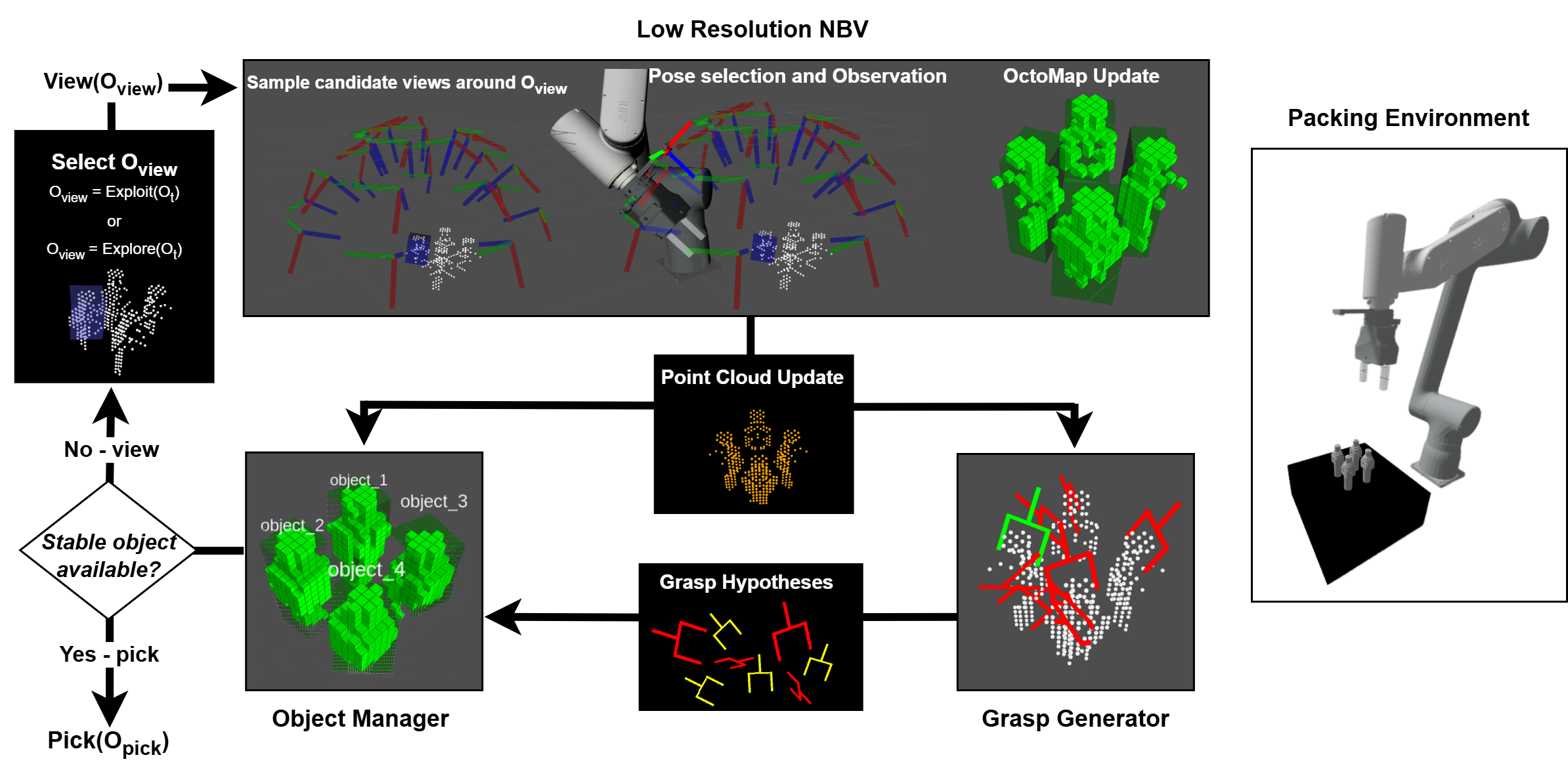}    
\caption{Pipeline overview. The Object Manager selects \emph{pick} or \emph{view}. For \emph{view}, a target object is chosen (Exploration/Exploitation) and LR--NBV plans the next pose. The Grasp Generator outputs scored 6-DoF grasps. The loop repeats until all objects are removed.}
\label{fig:complete_pipeline}
\end{center}
\end{figure*}

\section{Related Work}

The task of selecting the viewpoint that maximizes information about a scene is known as the \emph{Next Best View} (NBV) problem. As denoted by~\cite{base_NBV}, NBV methods are commonly divided into \emph{model-based} and \emph{model-free} approaches. Model-based methods validate a known object model against observations, while model-free methods—our focus—assume no prior knowledge and iteratively reconstruct the scene. These pipelines typically rely on volumetric representations, such as probabilistic voxel grids (e.g., OctoMap), where space is discretized into free, occupied, and unknown regions. 
Given a set of sampled poses, the most informative one is typically selected through a utility function, often based on ray casting. These utilities quantify informativeness and can be adapted to both task objectives and hardware constraints. Most classical approaches (\cite{2_utility_volumetric}) focus on exploration gain, i.e., maximizing unobserved space. Several work incorporates hardware-aware terms, such as minimizing robot motion or favoring optimal sensor incidence angles (\cite{4_utility}). However, these formulations are not well suited to low-resolution sensing. Sparse observations make nearby viewpoints hard to distinguish, leading to similar utility scores and redundant acquisitions. While redundancy can be mitigated through penalties on revisited areas as~\cite{reference_paper}, such strategies are less effective in object-centric scenarios, where local aliasing between neighboring poses remains a challenge. To address this, we prioritize under-sampled regions while reducing overlap with already observed volumes.

Beyond reconstruction-driven NBV, recent work integrates grasp planning into view selection. \cite{grasp_heuristic} rank candidate grasps using a view database and alignment heuristics, but their reliance on prior data limits generalization and can bias exploration under partial observations. Similarly, \cite{grasp_multi_fing} refine geometry near predicted contacts for multi-finger grasping, requiring accurate local surface information that is difficult to obtain with low-resolution sensing. In cluttered environments, \cite{grasp_clutter} combine pixel-wise grasp proposals with information gain, though their effectiveness relies on high-quality RGB-D data.
More closely related, \cite{grasp_stability} propose a closed-loop NBV strategy that iteratively refines grasp hypotheses until a stable grasp is found. This stability criterion remains appealing under low-resolution sensing, as grasp quality can improve with additional views. We adopt a similar principle but extend it to multi-object packing, introducing an object-centric scheduler that determines which item to observe or pick. Unlike \cite{grasp_stability}, which assumes known object properties, our method estimates object size and pose online from minimal initial observations and updates them as new data become available.

\section{Method Overview}
\label{Method Overview}
We aim to minimize the number of acquisitions---and thus the makespan---by deciding at each iteration whether to refine an object reconstruction (\emph{view}) or to execute a grasp (\emph{pick}). This decision is taken by the Object Manager (Sec. \ref{subsec:Object_Manager}) based on two inputs: the current geometric state and the grasp hypotheses. The former is represented by an OctoMap \(M_t\) and its associated point cloud \(P_t\), both updated whenever a new observation arrives. If \emph{view} is chosen, LR--NBV (Sec.~\ref{Low Resolution Next Best View}) selects the next pose \(x_T\in \mathrm{SE}(3)\); the sensor at \(x_T\) returns \(D(x_T)=\{p_i\}_{i=1}^{R}\subseteq\mathbb{R}^3\), where \(p_i\) are the 3D points measured at pose \(x_T\) and \(R\) is set by the sensor resolution. The observation is fused to update \(M_t=\mathcal{F}(M_{t-1},D(x_T))\) and the derived \(P_t\), and thus the per-object reconstructions extracted from \(M_t\). In parallel, at every map update, the Grasp Generator (Sec.~\ref{Grasp Generator}) takes as input the latest \(P_t\) to propose and score candidate grasps; these cues feed back into the Object Manager to arbitrate the pick--vs--view decision at the next iteration. Fig.~\ref{fig:complete_pipeline} depicts the flow and module interactions.

\subsection{Grasp Generator}
\label{Grasp Generator}
Whenever the occupancy map \(M_t\) is updated with a new acquisition \(D(x_T)\), its point cloud \(P_t\) is fed to this module, which returns a list of candidate grasps:
\[
G_t \;=\; \mathcal{G}(P_t) \;=\; \{\, g_{t,1},\ldots,g_{t,N_t} \,\}, 
\qquad g_{t,m} = (T_{t,m}, q_{t,m}),
\]
where each grasp \(g_{t,m}\) consists of a 6-DoF pose \(T_{t,m}\in \mathrm{SE}(3)\) and an associated quality score \(q_{t,m}\in\mathbb{R}\).
The Grasp Generator operates directly on the point cloud and can be either purely geometric or learning-based; in both cases, it estimates plausible gripper poses and assigns a score that reflects grasp stability and feasibility (e.g., antipodal consistency, collision risk via \(M_t\), and reachability).
This procedure is executed at every map update so that \(G_t\) stays aligned with the evolving scene.

\subsection{Object Manager}
\label{subsec:Object_Manager}
The Object Manager fuses geometric evidence and grasp hypotheses across iterations to decide \emph{what to view} and \emph{when to pick}. 
Its guiding idea is \emph{stability by re-observation}: in low-resolution settings, early reconstructions may be partial and induce grasps that reflect only local geometry and are thus unreliable. 
By monitoring how grasp proposals evolve as the map improves, the module favors grasps that reappear consistently over time and treats them as more trustworthy.
To evaluate this stability and arbitrate between further reconstruction (view) and grasping (pick), the Object Manager processes the raw point cloud \(P_t\) from the occupancy map \(M_t\) and the \emph{global} grasp set produced by the Grasp Generator.
With that information, it maintains the scene objects as:
\[
\mathcal{O}_t=\{o_t^{(1)},\ldots,o_t^{(N_t)}\},\;
o_t^{(n)}=\big(B_t^{(n)},\,\mathcal{G}^{\mathrm{obj}}_t(o_t^{(n)}),\,S_t(o_t^{(n)})\big)
\]
where \(B_t^{(n)}\) is the oriented bounding box (extent and pose),
\(\mathcal{G}^{\mathrm{obj}}_t(o_t^{(n)})\) is the per-object grasp list for \(o_t^{(n)}\),
with generic entry \((T_{o,j},\, q_{o,j},\, c_{o,j},\, \hat s_{o,j})\).
The first two terms \((T_{o,j}, q_{o,j})\) are as introduced in Sec.~\ref{Grasp Generator};
\(c_{o,j}\in\mathbb{R}\) is an observation counter tracking re-occurrences of equivalent grasps;
and the grasp score \(\hat s_{o,j}\in\mathbb{R}\) accounts for both the current quality and the counter,
with the specific update detailed in Alg.~\ref{alg:grasp-while}.
Finally, \(S_t(o_t^{(n)})\in\mathbb{R}\) denotes the object-level stability used for the pick/view decision.
The module comprises two stages:
Object Identification, and Pick-or-View Arbitration.

Object identification is performed directly from the raw point cloud \(P_t\), without prior knowledge of object pose or size. The occupancy map \(M_t\) is updated by classifying voxels as free, occupied, or uncertain, enabling segmentation of the scene.
We apply DBSCAN by~\cite{DBSCAN} to cluster points based on density. Each cluster defines an object hypothesis, enclosed in an oriented bounding box (OBB). The OBB encodes object extent, defines a dynamic region of interest (ROI) for view planning, and supports collision checking.
Temporal consistency is enforced by matching objects in \(\mathcal{O}_t\) and \(\mathcal{O}_{t-1}\) using centroid displacement and bounding-box overlap, allowing tracking, detection of new objects, and handling of occlusions.

The pick-or-view arbitration performs the core decision of the Object Manager. It first collects representative object poses \emph{(line~2)} and associates each incoming grasp \((T_i, q_i)\) to the nearest object via a pose--distance criterion, then checks for matches with previously stored grasps \emph{(lines~3--6)}. If a match is found, the observation counter \(c\) and stability \(\hat{s}\) are updated \emph{(lines~8--9)}; otherwise, a new entry is added \emph{(line~11)}.
Object-level stability is then computed by aggregating grasp scores \emph{(lines~14--15)}. If an object stability exceeds \(\tau_{\text{stab}}\), it is selected for picking \emph{(lines~17--20)}; otherwise, a view action is triggered \emph{(lines~23--27)}.
In this case, the target object \(O_{\text{view}}\) is selected by \(\pi_{\text{view}}\): \emph{Exploration} chooses the least-stable object, while \emph{Exploitation} selects the most-stable one, and \(\textsc{View}(O_{\text{view}})\) is executed \emph{(line~27)} via LR--NBV (Sec.~\ref{Low Resolution Next Best View}).

\begin{algorithm}[!b]
\scriptsize
\caption{Pick Or View}
\label{alg:grasp-while}
\begin{algorithmic}[1]
\Require $\mathcal{O}_t$, $G_t$, thresholds $\tau_d,\tau_g,\tau_{\text{stab}}$
\Statex \hspace{\algorithmicindent}view policy $\pi_{\text{view}}\in\{\textsc{Explore},\textsc{Exploit}\}$
\While{$|\mathcal{O}_t| > 0$}
  \State $\mathbf{T}^{\text{obj}} \gets [\,\bar T_t(o)\ \text{for}\ o\in\mathcal{O}_t\,]$
  \ForAll{$(T_i,q_i)\in G_t$}
    \State $n^\star \gets \textsc{FindNearest}(T_i,\mathbf{T}^{\text{obj}};\tau_d)$
    \State $\mathbf{T}^{\text{grasp}} \gets [\,T_{n^\star,j}\ \text{for}\ (T_{n^\star,j},\cdot)\in \mathcal{G}^{\mathrm{obj}}_t(o_t^{(n^\star)})\,]$
    \State $j^\star \gets \textsc{FindNearest}(T_i,\mathbf{T}^{\text{grasp}};\tau_g)$
    \If{$j^\star \neq \texttt{none}$}
      \State $c_{n^\star,j^\star} \gets c_{n^\star,j^\star}+1$
      \State $\hat{s}_{n^\star,j^\star} \gets q_{n^\star,j^\star}\,c_{n^\star,j^\star}$
    \Else
      \State add $(T_i,\ q_i,\ c{=}1,\ \hat{s}{=}q_i)$ to $\mathcal{G}^{\mathrm{obj}}_t(o_t^{(n^\star)})$
    \EndIf
  \EndFor
  \ForAll{$o \in \mathcal{O}_t$}
    \State $S_t(o) \gets \sum_{(T,q,c,\hat{s}) \in \mathcal{G}^{\mathrm{obj}}_t(o)} \hat{s}$
  \EndFor
  \State $\mathcal{R}_t \gets \{\, o \in \mathcal{O}_t \mid S_t(o) \ge \tau_{\text{stab}} \,\}$
  \If{$\mathcal{R}_t \neq \emptyset$}
    \State $O_{\text{pick}} \gets \arg\max_{o \in \mathcal{R}_t} S_t(o)$
    \State \textsc{Pick}$(O_{\text{pick}})$
  \Else
    \If{$\pi_{\text{view}}=\textsc{Explore}$}
      \State $O_{\text{view}} \gets \arg\min_{o\in\mathcal{O}_t} S_t(o)$
    \Else
      \State $O_{\text{view}} \gets \arg\max_{o\in\mathcal{O}_t} S_t(o)$
    \EndIf
    \State \textsc{View}$(O_{\text{view}})$
  \EndIf
\EndWhile
\end{algorithmic}
\end{algorithm}

\subsection{Low Resolution Next Best View}
\label{Low Resolution Next Best View}
This module describes the \emph{view} action used in Alg.~\ref{alg:grasp-while} \emph{(line~27)} by the Object Manager. Given the view target \(O_{\text{view}}\), the view--planning module evaluates candidate sensor poses \(x_T\) to decide the next acquisition. Candidates are sampled on a viewing sphere centered at \(O_{\text{view}}\).
The selection among candidate poses is guided by the LR--NBV utility introduced in our prior work \citet{preziosa2025low}, which we report here for completeness.
The utility is tailored for low-resolution sensing: it balances exploration of uncertain space with a preference for sparse, under-represented regions, thereby discouraging revisits.
Formally, it is a weighted sum of two gains—the Exploration gain \((G_e)\) and the Density gain \((G_d)\):

\vspace{-1.2em}
\begin{equation}
U(x_n) = w_e G_e(x_T) + w_d G_d(x_T)
\label{eq:utility}
\end{equation}
\vspace{-1.2em}

with \( w_e, w_d >0 \) indicating the per-gain weights; these are adjustable to balance the contributions of each term for a given sensing modality.
\begin{figure}
\begin{center}
\includegraphics[width=8.0cm]{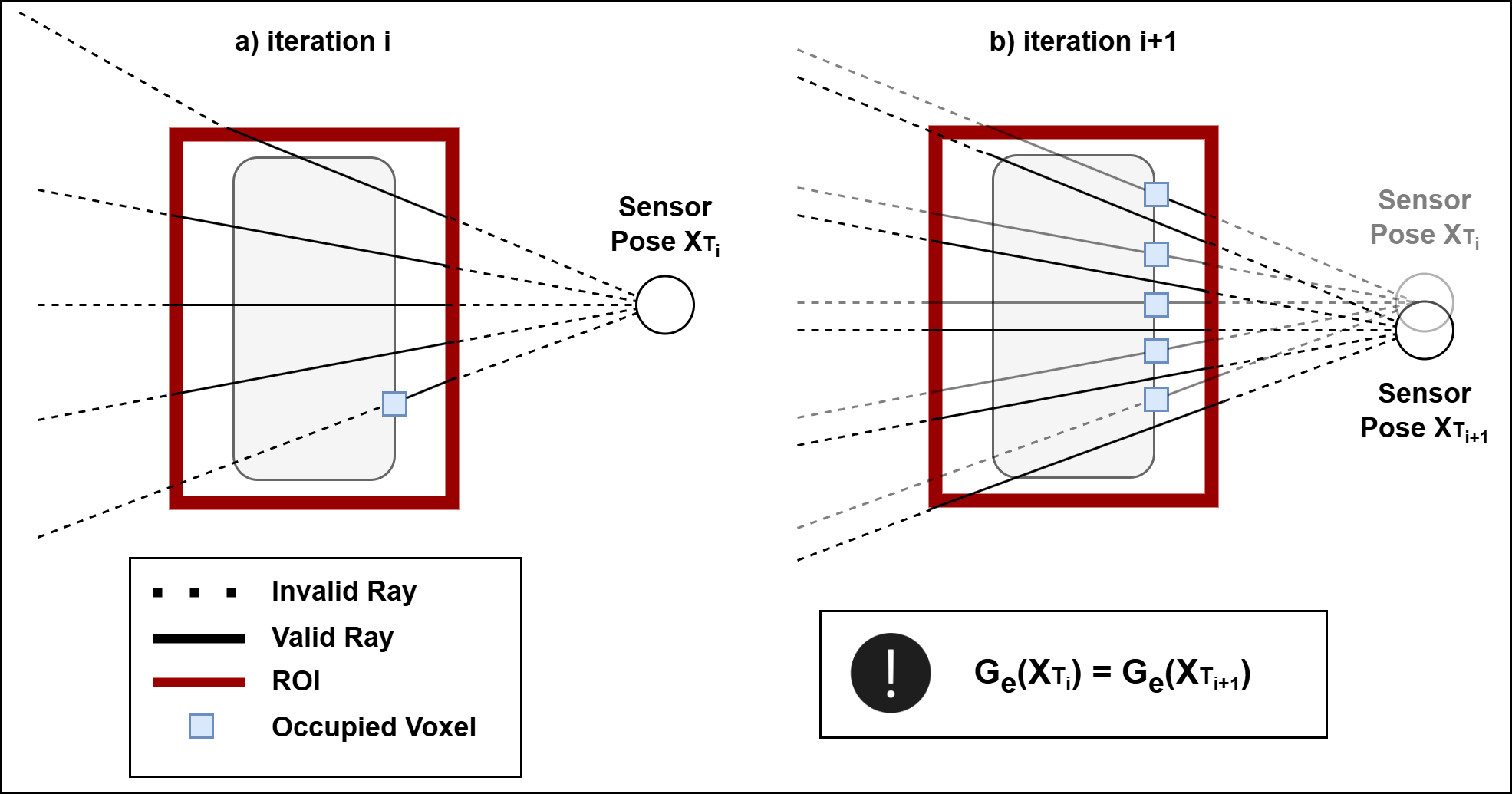}    
\caption{Ray validity and Exploration gain. Low resolution leads to similar scores and redundant views. Image from~\cite{preziosa2025low}.}
\label{fig:ray_validity}
\end{center}
\end{figure}
Both gains rely on a ray-casting procedure that emulates the sensor at a candidate pose. Each ray \( \boldsymbol{r} \in \mathcal{R}(x_n, \text{FoV}, \text{res}) \) originates from the sensor pose \( x_T \) and is given by \( \boldsymbol{r}(t) = \mathbf{o} + t \mathbf{d} \),
where \( \mathbf{o} \in \mathbb{R}^3 \) is the origin, \( \mathbf{d} \in \mathbb{R}^3 \) the direction, and \( t \) the distance along the ray. Let \( t^* \) be the first point where \( \boldsymbol{r}(t) \) intersects the object \(o_t^{(i)} \), we define ray validity as:
\begin{equation} \label{eq:ray_validity}
\begin{small}
V(\bold{r}, t) =
\begin{cases} 
1 & \text{if } \bold{r}(t) \cap \text{ROI} \neq \emptyset \text{ and } \exists t^* \leq t : \bold{r}(t^*) \in o_t^{(i)}, \\
0 & \text{otherwise.}
\end{cases}
\end{small}
\end{equation}
As illustrated in Fig.~\ref{fig:ray_validity}a, a ray segment is valid only within the ROI and when not occluded by the object, reflecting that regions beyond occlusions are not observable; thus only visible workspace portions contribute to reconstruction.
With this, the Exploration gain \( G_e(x_T) \) is:
\begin{equation} \label{eq:exploration_gain}
G_e(x_T) = \sum_{\boldsymbol{r} \in \mathcal{R}(x_T, \text{FoV}, \text{res})} \sum_{t_k = 0}^{1} V(\boldsymbol{r}, t_k) W(\boldsymbol{r}, t_k)
\end{equation}
where \( t_k \) are uniformly spaced samples between \( t = 0 \) and \( t = 1 \). The term \( W(\boldsymbol{r}, t_k) \) measures the contribution of each segment to volumetric exploration with traversed voxel volume as in~\cite{2_utility_volumetric}. Combining validity and volumetric terms focuses acquisitions on high-uncertainty areas, maximizing novel information.
However, as shown in Fig.~\ref{fig:ray_validity}b, under low-resolution sensing, nearby poses can yield identical \(G_e\). Since \(G_e\) ignores local point density, it may promote redundant views. To address this, we define the Density gain:
\begin{equation} \label{eq:density_gain}
G_d(x_T) = \sum_{\boldsymbol{r} \in \mathcal{R}(x_T, \text{FoV}, \text{res})} \sum_{t_k = 0}^{1} V(\boldsymbol{r}, t_k) \cdot \frac{1}{1 + \rho(\boldsymbol{r}, t_k)}
\end{equation}
Here, \( \rho(\boldsymbol{r}, t_k) \) is the number of object points within a spherical neighborhood of radius \( r_d \) centered at \( \boldsymbol{r}(t_k) \).
This downweights rays through dense areas and prioritizes under-sampled regions, thereby reducing redundancy.

\FloatBarrier
\section{Experimental Validation}
\label{sec:experiments}
We experimentally validate two aspects of the method. Study~A (Sec.~\ref{Study A}) assesses the effectiveness of the proposed low-resolution utility in selecting informative poses under severe sensing constraints. Study~B (Sec.~\ref{Study B}) evaluates the behavior of the full pipeline with the Object Manager—i.e., when to keep reconstructing or pick.

\begin{figure*}[h]
\begin{center}
\begin{tabular}{cc}
{\hspace{-0.4cm}\includegraphics[width=0.48\textwidth]{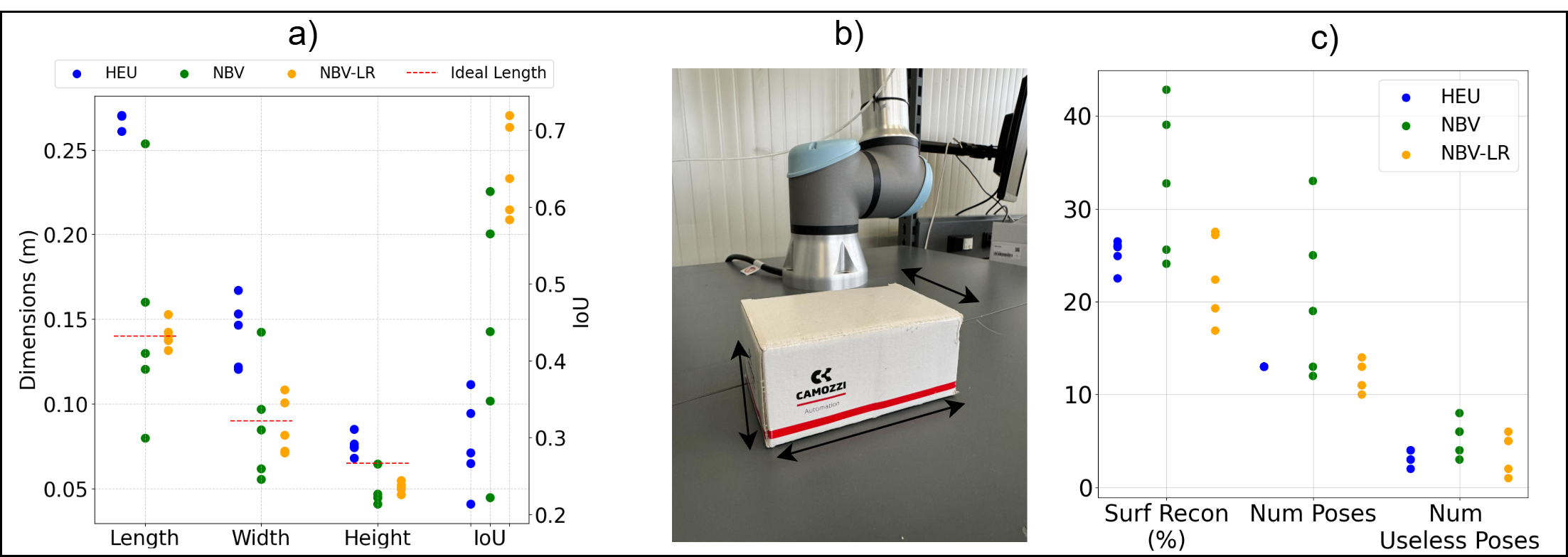}} & 
{\hspace{-0.1cm}\includegraphics[width=0.48\textwidth]{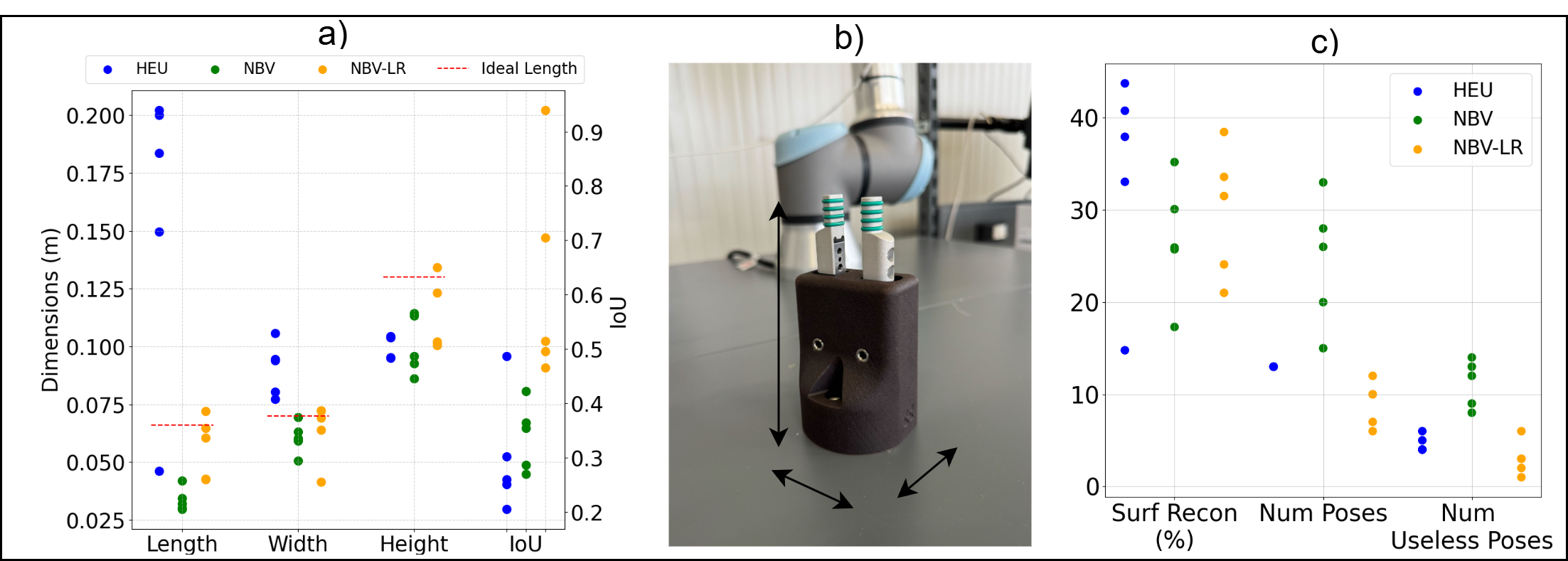}}  \\ 
{\hspace{-0.4cm}\includegraphics[width=0.48\textwidth]{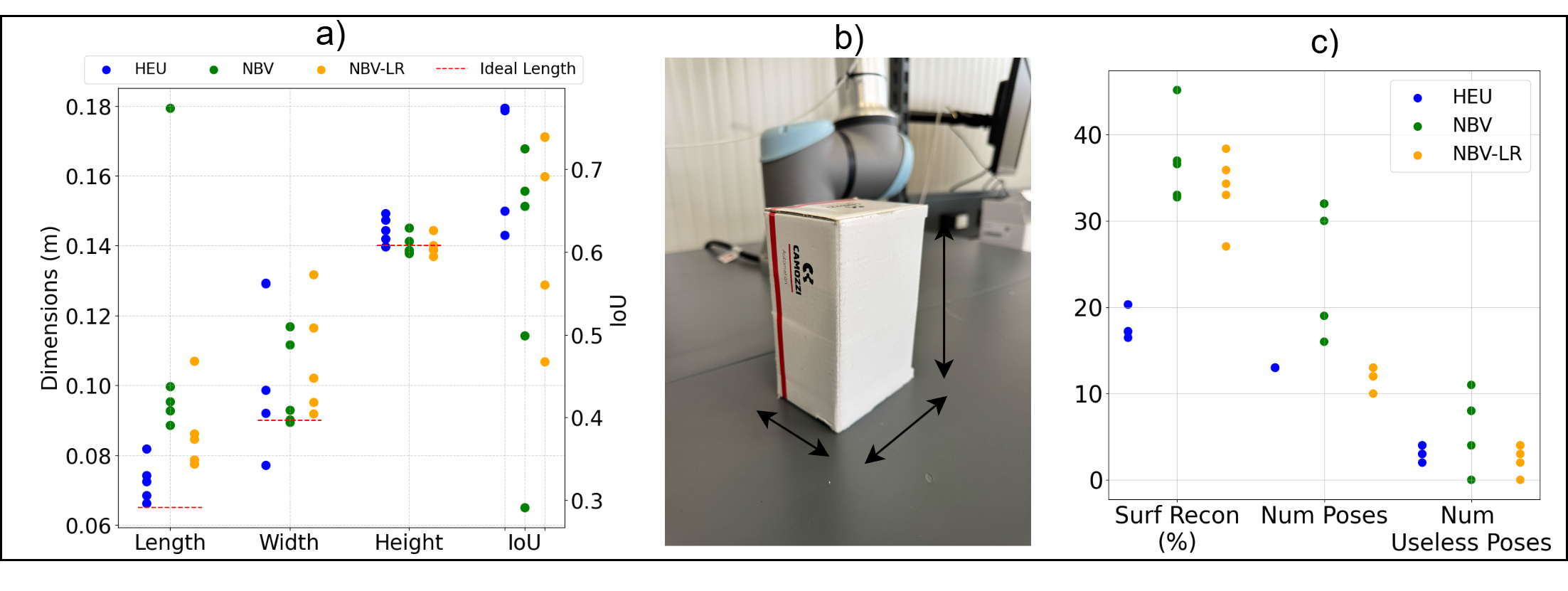}} & 
{\hspace{-0.1cm}\raisebox{0.17cm}{\includegraphics[width=0.48\textwidth]{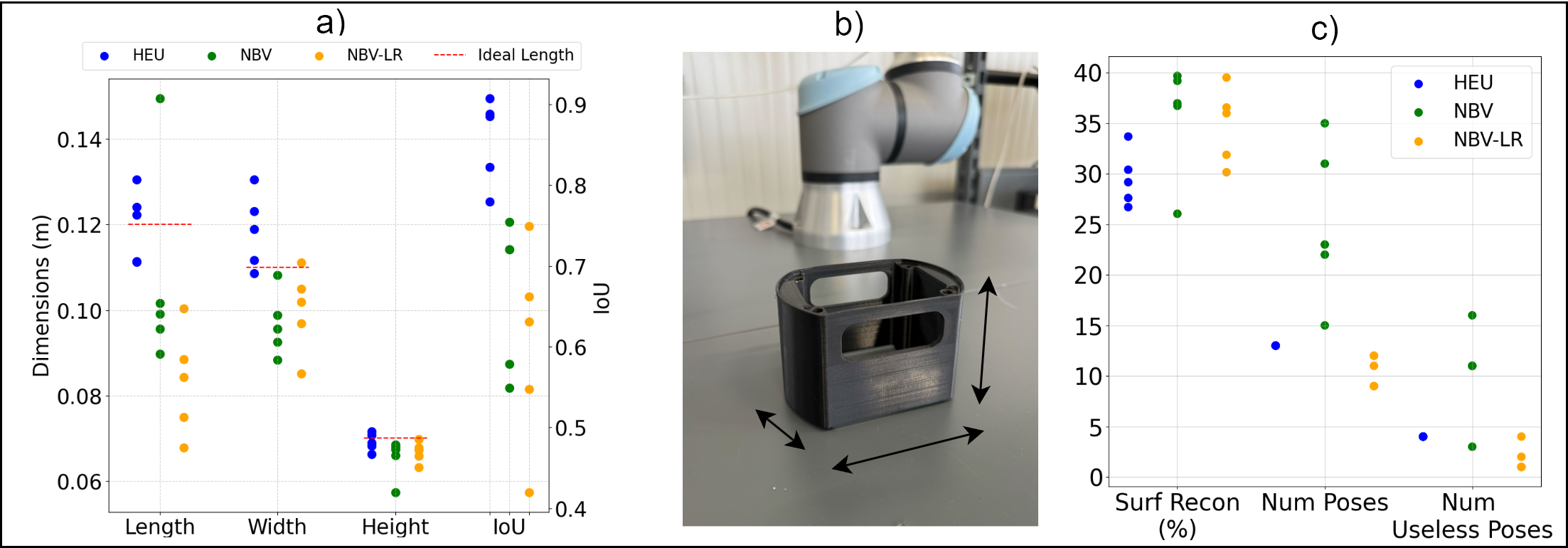}}} \\ 
\end{tabular}
\caption{Acquisition results for four objects: horizontal box (top left), vertical box (bottom left), gripper (top right), and flange (bottom right). For each object: (a) estimated dimensions and IoU; (b) canonical dimensions; (c) surface coverage, number of poses, and useless poses. Image from \cite{preziosa2025low}.}
\label{fig:all_experiments}
\end{center}
\end{figure*}

\subsection{Study A: Utility Function Validation}
\label{Study A}
This study builds on our previous validation (\cite{preziosa2025low}).
We compare LR-NBV with a standard \emph{NBV} method that uses the same \emph{Exploration gain} (Sec.~\ref{Low Resolution Next Best View}) but omits \emph{Density gain}, reflecting typical reconstruction-oriented NBV methods.
Since the experiment targets \emph{reconstruction only}, both methods use the same marginal-gain stopping criterion: at iteration \(t\), for pose \(x_{T_t}\) with utility \(U_t\), the marginal gain \(\Delta_t = U_t(x_{T_t}) - U_{t-1}(x_{T_{t-1}})\) is computed, and acquisition stops when \(\Delta_t \le \varepsilon_{\text{mg}}\).
As a second reference, we include a heuristic (\emph{HEU}) that samples 13 viewpoints along an ellipse centered on the workspace, assuming known object pose.
Each of the four objects in Fig.~\ref{fig:all_experiments} is reconstructed five times with all methods.

\begin{table}[hb]
\centering
\caption{Study A parameters.}
\label{tb:parameters}
\begin{tabular}{cccc}
\hline
sampled\_poses & \(w_e\) & \(w_d\) & OctoMap\_res[m] \\ \hline
50 & 500 & 0.05 & 0.01 \\ \hline
\end{tabular}
\end{table}

Experiments were carried out on a real UR5e manipulator equipped with a VL53L8CX time-of-flight sensor mounted on the end effector. The device has a $45^\circ \times 45^\circ$ field of view and an $8{\times}8$ pixel array, yielding extremely sparse observations. Computation ran on a Lenovo Yoga Pro 7i Gen 9 (Intel\textsuperscript{\textregistered} Core\texttrademark~i7-13700H, 32\,GB RAM). Table~\ref{tb:parameters} reports the parameter values used in all experiments.
We evaluate performance using four metrics: (i) bounding-box dimensions with intersection-over-union (IoU) to quantify reconstruction accuracy—relevant for container-size selection in packing lines; (ii) number of poses, used as a proxy for time efficiency in packing scenarios; (iii) reconstructed surface coverage; and (iv) count of useless poses, namely viewpoints that fail to increase coverage and may temporarily degrade reconstruction quality. Fig.~\ref{fig:all_experiments} summarizes the metrics over five runs per object.

We applied the Mann–Whitney U test to compare \emph{LR-NBV}, \emph{NBV}, and \emph{HEU} across all metrics and objects. \emph{LR-NBV} achieves statistically significant gains ($p{<}0.05$) in IoU for \emph{horizontal box} and \emph{gripper}. For \emph{vertical box} and \emph{flange}, significant differences ($p{<}0.05$) appear in at least one bounding-box dimension (length, width, or height), indicating more accurate size estimates in several cases relative to the baselines. In terms of efficiency, \emph{LR-NBV} consistently requires fewer poses ($p{<}0.05$)—often less than half of \emph{NBV}—while maintaining comparable reconstruction percentages. This efficiency stems from better prioritization of informative viewpoints, reducing redundancy. Conversely, \emph{NBV} exhibits more \emph{useless poses}, reducing coverage and requiring extra views.

\subsection{Study B: Full Pipeline Evaluation}
\label{Study B}
In this study we evaluate the end-to-end robotic packing pipeline (Sec. \ref{Method Overview}) under realistic operating conditions. We compare the two view policy (Sec.~\ref{Grasp Generator}): \emph{Exploration} \emph{XPLR}, which at each step selects the cluster with the \emph{lowest} stability score to spread sensing across objects, and \emph{Exploitation} \emph{XPLT}, which selects the cluster with the \emph{highest} stability score to focus on a specific object and trigger an early grasp. The two types of objects used in the experiments are shown in Fig.~\ref{fig:makespan_exp}, which are actual production parts used in a packing line. To isolate sensing constraints, we evaluate two effective input resolutions: {R\textsubscript{60}} (60$\times$60) and {R\textsubscript{30}} (30$\times$30).

Experiments were conducted on an \emph{ABB GoFa 12} collaborative arm with a \emph{MaixSense A010} time-of-flight sensor mounted on the end effector ($70^\circ \times 60^\circ$ FoV, $100{\times}100$ resolution). Inputs were downsampled to the effective resolutions \mbox{$R_{60}$} and \mbox{$R_{30}$}. 
Grasp generation is performed by \emph{GraspNet} (\cite{graspnet}) in a depth-only configuration, producing 6D grasp hypotheses with associated scores from the point cloud. 
In each episode, three objects in random poses are provided on a pallet, and the robot transfers them sequentially into a container; a single initial randomized acquisition is adopted.
We report: (i) grasp success rate (\%), (ii) mean surface reconstruction (\% coverage) at grasp time, (iii) failure breakdown by cause, and (iv) makespan to empty the pallet with the required acquisitions to grasp each object. 
Policy and resolution define a \(2\times2\) design (R\textsubscript{60}/R\textsubscript{30}, XPLT/XPLR), evaluated on two object types. For each setting, we perform 16 trials (8 per object), yielding 64 trials and \(T=192\) grasp attempts, with no re-attempts. The experiment continues until the pallet is cleared.

\begin{table}[tbp]
\begin{center}
\caption{Grasp success}\vspace{-5pt}\label{tb:grasp_success} 
\begin{tabular}{ccccc}
Object & R60--XPLR & R60--XPLT & R30--XPLR & R30--XPLT \\ \hline
Obj~A  &  62.5\% &  \textbf{70.8\%} &  54.2\% &  54.2\% \\
Obj~B  &  66.7\% &  58.3\% &  \textbf{83.3\%} &  75.0\% \\ \hline
\end{tabular}
\end{center}
\end{table}

\subsubsection{Grasp success}
We executed $T=192$ grasp attempts overall: $S=126$ successes ($65.6\%$) and $F=66$ failures ($34.4\%$). 
Table~\ref{tb:grasp_success} reports success rate over 24 attempts for each approach. Reading the table in light of the object properties visible in Fig.~\ref{fig:makespan_exp}—Obj~A is much larger, whereas Obj~B is smaller and includes reflective areas—two trends emerge. First, higher resolution (\mbox{$R_{60}$}) makes large parts easier to reconstruct and grasp: denser, more coherent geometry stabilizes pose estimation and improves grasp ranking, explaining better success on Obj~A. Second, for small, shiny parts, \mbox{$R_{30}$} often outperforms \mbox{$R_{60}$}: higher resolution captures more false reflections from reflective surfaces, adding spurious points that confuse clustering; the lower resolution acts like mild spatial smoothing, reducing noise and yielding more reliable grasps on Obj~B.

\begin{table}[hb]
\begin{center}
\caption{Mean reconstruction at grasp~time} \vspace{-5pt}\label{tb:recon_means}
\begin{tabular}{ccccc}
Object & R60--XPLR & R60--XPLT & R30--XPLR & R30--XPLT \\\hline
Obj~A  & 42.55\% & \textbf{56.29\%} & 39.79\% & 44.36\% \\
Obj~B  & 54.77\% & \textbf{59.83\%} & 50.45\% & 47.32\% \\ \hline
\end{tabular}
\end{center}
\end{table}

\subsubsection{Surface reconstruction}
At each grasp trial, we measured the surface reconstruction at the decision point (i.e., when the object was judged grasp-ready). The means in Table~\ref{tb:recon_means} lie consistently in the \(40\text{–}60\%\) range across all policy–resolution pairs and both objects. This aligns with our design goal: the pipeline does not seek exhaustive geometry but a partial, grasp-ready model. At the same time it avoids under-reconstructed cases (below 40\%) where grasp hypotheses are possible but unstable.

\begin{figure}
\begin{center}
\includegraphics[width=8cm]{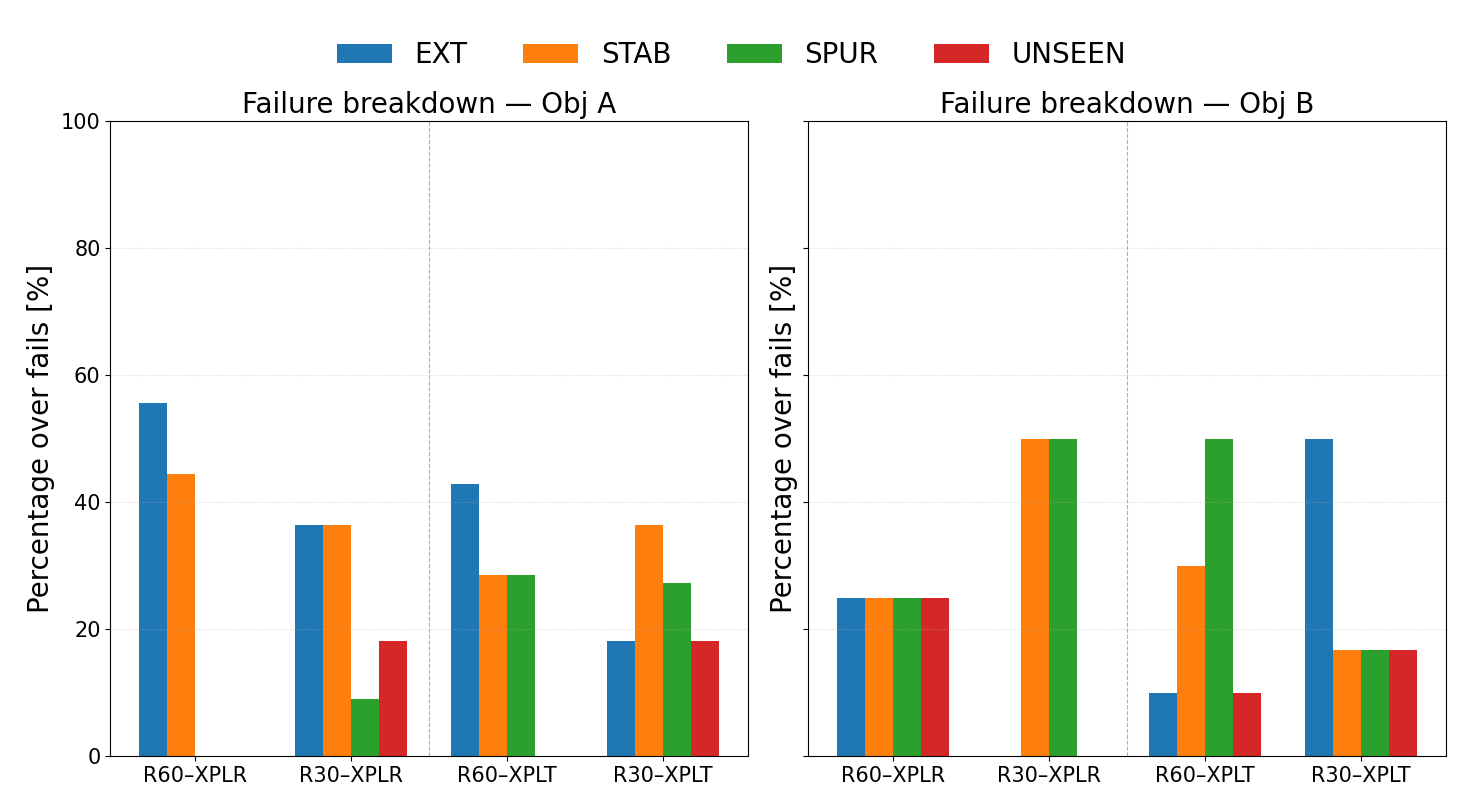}    
\caption{Failure breakdown by object and condition (percentages over fails)} 
\label{fig:fails}
\end{center}
\end{figure}

\subsubsection{Grasp failures}
To interpret the coverage results, failures are classified by cause: 
i) \emph{STAB} (stability–readiness failure): we deem an object grasp–ready when one or more grasp hypotheses persist across iterations. We classify a failure as STAB when the object was grasp–ready and the proposed grasp is consistent with the locally reconstructed patch, yet execution fails because unreconstructed regions are essential for grasp success.
ii) \emph{SPUR}: grasps are proposed on spurious points caused by sensor noise, often amplified by reflective surfaces; 
iii) \emph{UNSEEN}: the object is never sufficiently observed or not segmented.
iv) \emph{EXT}: failures due to external constraints (e.g., motion planning or workspace limitations).


Across all conditions we observed $F=66$ failures. Percentages in Fig.~\ref{fig:fails} are computed \emph{within the failures of each condition}. Focusing on STAB, it accounts for $33.3\%$ of all failures (weighted across conditions). By policy, XPLR averages $37.5\%$ STAB whereas XPLT averages $29.4\%$, indicating that XPLT reduces false positives by enforcing a stricter, object-specific verification of grasp readiness. Despite this, the stability index under low-resolution reconstruction remains effective, as reflected by the overall grasp success.
The breakdown indicates that noise from low-cost, low-resolution sensing is a major error source, strongly influenced by reflectivity and ambient light.

\begin{figure*}[h]
\begin{center}
\includegraphics[width=16.0cm]{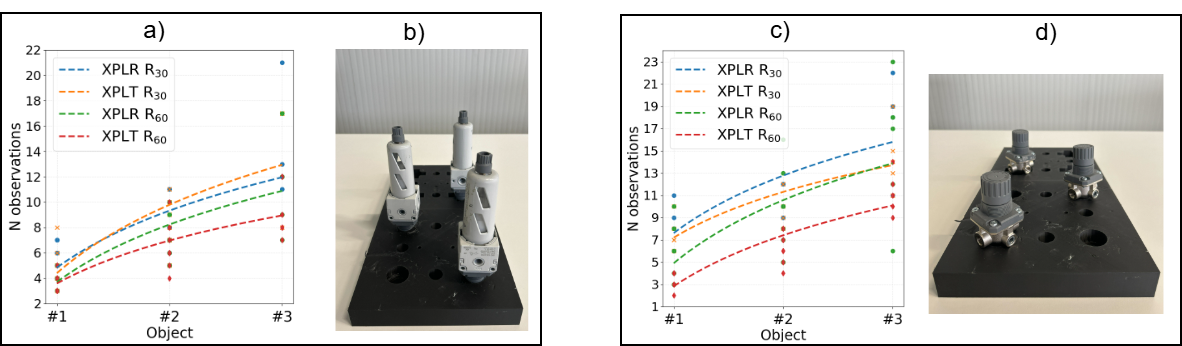}    
\caption{(a,c) Number of acquisitions required before each grasp for three objects, under different acquisition and resolution modes. The x-axis denotes the grasp index (\#1–\#3). (b,d) Corresponding experimental setups for Obj~A and Obj~B.}
\label{fig:makespan_exp}
\end{center}
\end{figure*}

\subsubsection{Acquisition curves}
Fig.~\ref{fig:makespan_exp}a and Fig.~\ref{fig:makespan_exp}c complement this view by showing the \emph{number of observations} required per object before grasp. In both figures, the two \mbox{$R_{30}$} curves lie above the corresponding \mbox{$R_{60}$} curves—consistent with the fact that higher resolution captures more informative points per view and thus needs fewer acquisitions to reach graspability. All curves exhibit a sublinear progression: after grasping Object~\#1, many earlier views already contributed evidence for the remaining items, so Objects~\#2–\#3 typically require progressively fewer additional observations. Policy-wise, \emph{XPLR} tracks a clearer logarithmic pattern (it lifts the least-reconstructed item at each step), while \emph{XPLT} is closer to linear because it focuses on a single object until grasp and may leave the next one starting from scratch. This explains the flatter decay of the depth curves noted in both figures; notably, with low-resolution sensing, reflective areas can intermittently corrupt reconstruction and trigger partial tear-down, so—even for the smallest object—the overall makespan can remain comparable to the largest one.


\section{Conclusion}
This paper shows that low-resolution, low-cost perception is a viable solution for robotic packing. We propose an object-centric pipeline that combines a low-resolution NBV strategy (LR--NBV) with grasp-driven reconstruction and a stability by re-observation criterion.
Experiments demonstrate that low-resolution sensing can reliably estimate object extents for packing, while the integration of reconstruction and grasp stability enables effective decisions on when to grasp without requiring full geometry.
Future work will address more challenging scenarios, including highly reflective objects, by identifying and efficiently sensing problematic regions.

\begin{ack}
\vspace{-0.2cm}
The authors acknowledge \emph{Camozzi Research Center} for providing robotic equipment, as well as for their support.
\end{ack}

\bibliography{ifacconf}             

@article{graspingPaper2,
  title={End-to-end learning to grasp via sampling from object point clouds},
  author={Alliegro, Antonio and Rudorfer, Martin and Frattin, Fabio and Leonardis, Ale{\v{s}} and Tommasi, Tatiana},
  journal={IEEE RAL},
  year={2022},
}

@article{graspingPaper3,
  title={Grasp pose detection in point clouds},
  author={Ten Pas, Andreas and Gualtieri, Marcus and Saenko, Kate and Platt, Robert},
  journal={The Int. Journal of Robotics Research},
  year={2017},
}

@article{Packing_rgb_data,
  title={Smart pack: online autonomous object-packing system using RGB-D sensor data},
  author={Hong, Young-Dae and Kim, Young-Joo and Lee, Ki-Baek},
  journal={Sensors},
  year={2020},
}

@article{Packing_just_ordering,
  title={Dense robotic packing of irregular and novel 3D objects},
  author={Wang, Fan and Hauser, Kris},
  journal={IEEE Trans. on Robotics},
  year={2021},
}

@inproceedings{Packing_just_ordering_offline,
  title={Stable bin packing of non-convex 3D objects with a robot manipulator},
  author={Wang, Fan and Hauser, Kris},
  booktitle={IEEE ICRA},
  year={2019},
}

@inproceedings{Packing_low_cost,
  title={Towards robust product packing with a minimalistic end-effector},
  author={Shome, Rahul and Tang, Wei N and Song, Changkyu and Mitash, Chaitanya and Kourtev, Hristiyan and Yu, Jingjin and Boularias, Abdeslam and Bekris, Kostas E},
  booktitle={IEEE ICRA},
  year={2019},
}

@article{base_NBV,
author = {Scott, William R. and Roth, Gerhard and Rivest, Jean-Fran\c{c}ois},
title = {View planning for automated three-dimensional object reconstruction and inspection},
year = {2003},
journal = {ACM Comput. Surv.},
}

@article{2_utility_volumetric,
  title={Efficient autonomous exploration planning of large-scale 3-d environments},
  author={Selin, Magnus and Tiger, Mattias and Duberg, Daniel and Heintz, Fredrik and Jensfelt, Patric},
  journal={IEEE RAL},
  year={2019},
}

@INPROCEEDINGS{4_utility,
  title={A best next view selection algorithm incorporating a quality criterion},
  author={Massios, Nikolaos A and Fisher, Robert B and others},
  booktitle = {Proc. British Machine Vision Conference},
  year={1998}
}

@ARTICLE{reference_paper,
  author={Naazare, Menaka and Rosas, Francisco Garcia and Schulz, Dirk},
  journal={IEEE RAL}, 
  title={Online Next-Best-View Planner for 3D-Exploration and Inspection With a Mobile Manipulator Robot}, 
  year={2022},
}

@INPROCEEDINGS{DBSCAN,
  author={Deng, Dingsheng},
  booktitle={Int. Forum Electr. Eng. Autom.},
  title={DBSCAN Clustering Algorithm Based on Density}, 
  year={2020},}

@INPROCEEDINGS{spots,
    author={Nicola Maria Ceriani and Giovanni Buizza Avanzini and Andrea Maria Zanchettin and Luca Bascetta and Paolo Rocco},
  booktitle={IEEE ICRA}, 
  title={Optimal placement of spots in distributed proximity sensors for safe human-robot interaction}, 
  year={2013},
}

@INPROCEEDINGS{grasp_multi_fing,
  author={Arruda, Ermano and Wyatt, Jeremy and Kopicki, Marek},
  booktitle={IEEE/RSJ IROS}, 
  title={Active vision for dexterous grasping of novel objects}, 
  year={2016},
}

@INPROCEEDINGS{grasp_stability,
  author={Breyer, Michel and Ott, Lionel and Siegwart, Roland and Chung, Jen Jen},
  booktitle={IEEE/RSJ IROS}, 
  title={Closed-Loop Next-Best-View Planning for Target-Driven Grasping}, 
  year={2022},
}

@INPROCEEDINGS{grasp_clutter,
  author={Morrison, Douglas and Corke, Peter and Leitner, Jürgen},
  booktitle={IEEE ICRA}, 
  title={Multi-View Picking: Next-best-view Reaching for Improved Grasping in Clutter}, 
  year={2019},
}

@INPROCEEDINGS{grasp_heuristic,
  author={Gualtieri, Marcus and Platt, Robert},
  booktitle={IEEE/RSJ IROS)}, 
  title={Viewpoint selection for grasp detection}, 
  year={2017},
}

@inproceedings{graspnet,
  title={GraspNet-1Billion: A Large-Scale Benchmark for General Object Grasping},
  author={Fang, Hao-Shu and Wang, Chenxi and Gou, Minghao and Lu, Cewu},
  booktitle={Proc. IEEE/CVF Conf. Comput. Vis. Pattern Recognit. (CVPR)},
  year={2020},
}

@article{preziosa2025low,
  title = {Low Resolution Next Best View for Robot Packing},
  journal = {IFAC-PapersOnLine},
  note = {14th IFAC Symposium on Robotics ROBOTICS},
  year = {2025},
  author = {Giuseppe Fabio Preziosa and Chiara Castellano and Andrea Maria Zanchettin and Marco Faroni and Paolo Rocco},
}
                                                   







\end{document}